\documentclass{article}

\usepackage[english]{babel}
\usepackage[utf8]{inputenc}
\usepackage[T1]{fontenc}    % font encoding+
\usepackage{booktabs}
\usepackage[a4paper]{geometry}
\usepackage[authoryear]{natbib}
\usepackage{amsmath}
\usepackage{amsfonts}
\usepackage{amssymb}
\usepackage{graphicx}
\usepackage{xcolor}
\usepackage[colorlinks=true, allcolors=blue]{hyperref}
\usepackage{soulutf8}
\usepackage{fancyvrb}
\usepackage{comment}
\usepackage{todonotes}

\newcommand{\set}[1]{\mathcal{#1}}
\newcommand{\R}{\mathbb{R}}

\newcommand{\vect}[1]{\mathbf{#1}}

\title{Through the Thicket: A Study of Number-Oriented LLMs derived from Random Forest Models}
\author{Michał Romaszewski\footnote{mromaszewski@iitis.pl}, Przemysław Sekuła, Przemysław Głomb\\Michał Cholewa, Katarzyna Kołodziej}
\date{Institute of Theoretical and Applied Informatics,\\Polish Academy of Sciences\\\small{Bałtycka 5, 44-100 Gliwice, Poland\\\url{www.iitis.pl}}}
\begin{document}
\maketitle

\begin{abstract}
Large Language Models (LLMs) have shown exceptional performance in text processing. Notably, LLMs can synthesize information from large datasets and explain their decisions similarly to human reasoning through a chain of thought (CoT). An emerging application of LLMs is the handling and interpreting of numerical data, where fine-tuning enhances their performance over basic inference methods. This paper proposes a novel approach to training LLMs using knowledge transfer from a random forest (RF) ensemble, leveraging its efficiency and accuracy. By converting RF decision paths into natural language statements, we generate outputs for LLM fine-tuning, enhancing the model's ability to classify and explain its decisions. Our method includes verifying these rules through established classification metrics, ensuring their correctness. We also examine the impact of preprocessing techniques on the representation of numerical data and their influence on classification accuracy and rule correctness. \end{abstract}

\section{Introduction}

%First paragraph:
Large Language Models (LLM) are deep neural networks based on transformer architecture \cite{vaswani2017attention} which demonstrated exceptional performance in various tasks such as question answering~\cite{yeo2023assessing}, solving problems requiring general knowledge~\cite{ge2024openagi}, code generation~\cite{chen2021evaluating} or time-series understanding and processing~\cite{zhang2024large}. A notable feature of LLMs is their ability to synthesize information from large datasets and explain their decisions in a way that is similar to human reasoning. In particular, they can generate a chain of thought (CoT)~\cite{wu2024usable}, presenting statements that support the model's conclusions. Consequently, LLM may be the first method to provide efficient explanations and insights across multiple specialized domains using a single general model. 

%Second paragraph
An emerging application of LLMs is handling and interpreting numerical data, such as in tabular data mining~\cite{fang2024large} or classification tasks~\cite{dinh2022lift}, both of which involve matching patterns in numerical features relative to specific classes. While a basic approach relies on the model's general knowledge for zero-shot, one-shot, or few-shot inference, yielding sometimes accurate results~\cite{hou2024large, zhao2023large}, fine-tuning the LLM~\cite{dinh2022lift} enhances its understanding of specific problems or datasets, leading to improved performance.
A more interesting scenario involves the LLM providing the label and the supporting statements, forming a chain of thought -- a sequence of prompts that build upon each other to guide the model's response. This approach enhances model explainability and often improves performance, particularly in arithmetic and symbolic tasks~\cite{wei2023chainofthought}. An example is explainable temporal reasoning (ETR)~\cite{yuan2023back}, where the model infers from graph-structured data, classifies event outcomes, and supports its decisions with logical statements. This capability leverages general and domain-specific knowledge, offering an explainable, expert-like experience.

%Third paragraph
In LLM classification with sequence-to-sequence models, prompt and output preparation is essential. Data preparation involves serializing data structures into textual prompts and outputs~\cite{fang2024large}. Prompt preparation is an area of active research, focusing on assessing its impact on the model performance~\cite{sui2024table}. 
Output generation is straightforward when the model only needs to return a label~\cite{dinh2022lift}. 
However, in tasks like time series forecasting~\cite{xue2023promptcast}, a query may require a more complicated format, such as a series of values with context.
For tasks requiring both label generation and explanations, Knowledge Graphs are often employed~\cite{pan2024unifying} for input and output generation. Data is converted into a graph structure, with nodes representing facts and edges representing relationships. Natural language statements are generated by sampling the graph with random walks. In temporal knowledge graph (TKG) forecasting~\cite{lee2023temporal, yuan2023back}, the model must also maintain the correct order of statements to represent historical events accurately.

Recently, tree-based data structures have been explored for generating data-describing statements for LLMs. For example, in~\cite{litree}, decision trees enhance prompts with logical rules sampled from the tree. \cite{ziems2023explaining} proposes using decision tree explanations to train LLMs for Network Intrusion Detection (NID), with human-assisted evaluation of outputs. On the other hand,~\cite {zhuang2024learning} introduces MetaTree - a transformer-based, decision tree model. In our opinion, this direction holds significant potential as the data structures captured by decision trees can be converted into a sequential chain of thought statements, providing an effective method for training LLMs.

%Fourth paragraph
We propose a novel approach to training large language models (LLMs) by transferring knowledge from a random forest (RF). The RF~\cite{breiman2001random} is renowned for its efficiency and accuracy. Our method leverages the ensemble nature of RF by sampling multiple trees for each input to produce outputs for LLM fine-tuning. This approach is motivated by the widespread availability of training datasets for classical models and the universal applicability of RF in classification tasks. Additionally, while RF decisions can be explained through methods such as 
Shapley additive explanations (SHAP)~\cite{lundberg2017unified} or tree visualisation, LLMs can classify and explain their decisions in natural language. The broad adoption of LLMs shows that natural language explanations may be more comprehensible to users.

Our approach leverages the fact that given a label from a random forest (RF), the decision of each tree can be converted into propositional logic statements. We propose sampling the RF by selecting a subset of trees and tracing their classification paths. These paths form statements that translate into natural language, explaining the returned labels. Such conditions are objectively verifiable, a crucial aspect of model performance evaluation~\cite{zhou2023instructionfollowing}. To verify these conditions, we propose a methodology to find a subset of the training set described by the conditions and assess its properties using classification performance metrics. We also examine the impact of three preprocessing techniques: integer normalization, verbal description of values, and relation encoding -- on the representation of numerical data in text inputs/outputs and their impact on classification accuracy and rule correctness. Our contributions are as follows:

\begin{enumerate}
    \item We propose a novel method for training LLMs by using a random forest ensemble to generate accurate labels and corresponding rule-based explanations.
    \item We investigate LLMs' comprehension of numerical data, proposing several pre-processing operations that significantly impact the correctness of LLMs' decision explanations.
    \item We present a study where an LLM-based model replaces a traditional classification model through knowledge transfer, utilizing random forest ensemble rules transformed into LLM outputs.
    \item  Our method generates explanations alongside labels, providing arguments to justify decisions and potentially improving model performance by employing the chain of thought (CoT) approach.
    \item We introduce a method to verify the correctness of rules generated by LLMs by applying them to the training set and assessing their impact using established classification quality measures, providing an objective assessment of LLM outputs beyond classification accuracy.
\end{enumerate}

\subsection{Related work}
\label{sec:related_work}
\paragraph{Mathematical Reasoning} 
Since their introduction, LLMs have aimed to effectively represent and process natural language, including the task of handling mathematical reasoning. A significant challenge involves the accurate identification and manipulation of signs and symbols, particularly numbers, within the text. Mathematical reasoning is a crucial area that demonstrates the potential of LLMs in extracting and utilizing quantitative information to solve numerical tasks. However, this field is still in its early stages. \cite{wei2023cmath}~evaluated several LLM models on the CMATH Dataset of elementary school math word problems, revealing that only GPT-4 achieved the required accuracy across all grade levels. Most models showed decreased performance accuracy with increasing arithmetic or reasoning complexity, indicating that LLMs still need to be capable of handling a wide range of numerical problems.

Scientific research on LLMs for mathematical reasoning focuses on developing prompting techniques and fine-tuning methodologies~\cite{ahn2024large}. \cite{imani2023mathprompter}~propose the MathPrompter model based on zero-shot CoT prompting to address arithmetic tasks by dividing them into simple intermediate steps and generating multiple methods (Python code and analytical expressions) to solve these steps, validating each step's correctness. Similarly, \cite{yamauchi2023lpml}~proposes a framework that integrates CoT prompting with an external Python REPL tool to correct CoT mistakes. Self consistency~\cite{wang2023selfconsistency} is a prompting method that utilizes multiple CoT reasoning paths, selecting the final answer by majority voting. A recent example of a fine-tuning-based method is the MetaMath model~\cite{yu2024metamath}, which bootstraps mathematical questions in the training dataset by rephrasing them to reflect forward and backward reasoning. \cite{magister2023teaching}~propose improving arithmetic solving performance through knowledge distillation, fine-tuning a small LLM model with CoT generated by a larger teacher model.

One critical aspect of handling numbers in LLMs is tokenization. Dividing complex numbers into multiple tokens can hinder their accurate interpretation. Several tokenization methods have been developed. Leading LLM models, GPT-3.5 and GPT-4, use separate tokens for 1-, 2-,  and 3-digit representations, while LLaMa and PaLM use a one-digit tokenization scheme. It has been shown that a right-to-left tokenization scheme (enforced by comma separation of numbers) improves mathematical reasoning performance compared to the traditional left-to-right technique~\cite{singh2024tokenization}. A notable approach to addressing the tokenization issue is the number decomposition method~\cite{muffo2023evaluating}, which supports each number in the prompt with an automatically generated textual description.

\paragraph{Prompting techniques}
Formulating queries (prompting) is crucial for enhancing LLM performance, with two primary methods being zero-shot and few-shot learning \cite{brown2020language}.
Few-shot Chain of Thought (CoT) prompting~\cite{wei2023chainofthought} introduces intermediate reasoning steps in prompts, improving performance and providing reasoning as part of the model's output. Building on this, zero-shot CoT prompting~\cite{kojima2023large} uses a single prompt template, '\textit{Let's think step by step}', to encourage task-agnostic reasoning. The simplicity and effectiveness of CoT prompting have led to various extensions, such as sub-problems division -- decomposing problems into simpler sub-problems, external assistance -- using external knowledge sources or tools~\cite{yu2023better}, and CoTGenius -- enhancing initial CoT prompts with derived topics and detailed reasoning paths~\cite{cheng2024chainlm}.

Another recent proposal is complexity-based prompting~\cite{fu2023complexitybased}, which selects examples with complex reasoning chains to improve multi-step reasoning. Federated prompting~\cite{liu2023federated} addresses frequently asked questions with identical reasoning paths by federating synonymous questions and using Self-consistency with majority voting or CoT to ensure consistent and practical solutions.

\paragraph{XAI} The concept of AI system explainability has been explored even for classical ML black-box systems, including neural networks and kernel SVMs. In~\cite{hoffman2018metrics}, various approaches to providing explanations are discussed, highlighting the importance of helping users understand model operations. This research is increasingly critical for complex models like Deep Neural Networks and LLMs. \cite{dwivedi2023explainable}~provides a comprehensive survey of contemporary methods, from Shapley additive explanations (SHAP) to individual conditional expectation (ICE) and partial dependancy plot  (PDP), focusing on one or a few features. These methods are widely used in decision-making processes in areas such as recommender systems~\cite{gawde2024explainable}, medical data~\cite{van2022explainable}, pharmaceutical engineering~\cite{polzer2022validation}, where XAI verifies ML responses and classifications. In fields such as cybersecurity~\cite{srivastava2022xai}, the use of XAI is particularly challenging due to the adversarial nature of the domain.

Advancements in natural language processing and LLMs have introduced new tools for AI decision-making explanations through natural language. \cite{cambria2023survey}~outlines the advantages of natural language explanations, including comprehensiveness and increased trust in decision-making. Works such as~\cite{luo2024understanding} highlight  XAI's positive impact on improving LLM results. 

Another intensively researched area involves systems that interact with users; for instance,~\cite{yu2022interaction} proposes a two-step XAI generative framework for providing arguments for classification labelling, while~\cite{nguyen2023black} incorporates XAI into a conversational agent.

\section{Method}
%Our input consists of a dataset with a feature table $\vect{X}\in\R^{N \times F}$ and a label vector $\vect{y}\in\R^N$. 

Given a dataset $\vect{\set{X}}\in\R^{N\times F}$ with $N$ examples and $F$ features and a set of labels $\set{Y}$, our input data to train the model is a training set $\set{T}_\text{train}=\{(\vect{x}_i,y_i)\}_{i=1,\ldots,|\set{T}_\text{train}|}$ where an example
$\vect{x}_i\in\set{X}$ and its label $y_i\in\set{Y}$.
Our goal is to create an LLM classifier that labels the data and explains its decisions. 
To do this, we require textual input-output pairs suitable for processing with the sequence-to-sequence (S2S) LLM model. These inputs are created by encoding the feature vectors into a textual form, while outputs are generated by sampling a random forest classifier trained on the set $\set{T}_\text{train}$.
\subsection{Preprocessing and substitution of numerical values}
\label{sec:method:preprocessing}
As presented in Section~\ref{sec:related_work}, the format of prompts/outputs and the way the numbers, signs, and symbols are represented significantly impact the ability of LLMs to perform mathematical reasoning. Particularly important for this work is the textual representation of conditional statements in the form:
\begin{equation}
\underbrace{\text{petal length (cm)}}_{\text{feature name}}
\quad\underbrace{4.80}_{\text{feature value}} 
\quad
\underbrace{\text{ >}}_{\text{condition}} \quad
\underbrace{\text{4.75,}}_{\text{threshold value}}
\label{eq:dtree_rule}
\end{equation}
which describes a comparison of data features with arbitrary thresholds. Tokenization converts such statements, including the floating-point values, into a series of tokens; for example, the GPT-4 tokenizer generates three tokens for each of the values in the Eq~\eqref{eq:dtree_rule}. To facilitate LLM understanding and reasoning with such statements, we perform several preprocessing steps, including: 
\begin{enumerate}
    \item Integer normalisation (IN) - transforming floating-point values to fixed-point values with a limited number of digits,
    \item Verbal description of values (VD) - extending the value with a verbal description based on box and whisker plot (box plot)~\cite{tukey1977exploratory},
    \item Relation encoding (RE) - converting inequality signs to words.
\end{enumerate}
We will use the above shortcuts when describing network hyperparameters, e.g. IN$+$VD$+$RE denotes that all the options are enabled.

\paragraph{Integer normalisation (IN)} Tokenisation has a major impact on how LLM performs arithmetic operations~\cite{singh2024tokenization}. Since LLMs tend to have problems with understanding large numbers and high-precision floating-point values, in our approach, we transform them into 2-digit numerical values encoded by 1 token by GPT-4 tokenisation scheme or 2 tokens by models employing single-digit tokenisation like LLaMa. For every feature, its values are converted into integers from a given range $\langle r_{\text{min}}, r_{\text{max}}\rangle$. For the data table $\vect{X}\in\R^{N\times F}$, obtained by concatenating vectors in the training set $\set{T}_\text{train}$, the transformation process is described by the following equation:
\begin{equation*}
\vect{X}_{\text{scaled}} = \left\lfloor \frac{(\vect{X} - \vect{v}_{\text{min}})}{(\vect{v}_{\text{max}} - \vect{v}_{\text{min}})} \times (r_{\text{max}} - r_{\text{min}}) + r_{\text{min}} \right\rceil,
\end{equation*}

where \( \vect{X}_{\text{scaled}} \) represents the scaled and integer-encoded data and vectors $\vect{v}_{\text{min}}, \vect{v}_{\text{max}} \in \R^F$ contain the minimum/maximum value of each feature. Since the transformation is performed on the training set, encoded values are then clipped to $\langle r_{\text{min}}, r_{\text{max}}\rangle$. The range used in our experiments was $\langle 0, 99\rangle$, so e.g. the expression {petal length (cm): 4.80}' becomes '\emph{petal length (cm): 30}'. The operation is lossy, introducing estimation error into values of example features and thresholds.

\paragraph{Verbal description of values (VD)}
The goal of this step is to provide the LLM with an understanding of how the feature values of a given example compare to the entire training set. To support this, we augment numerical values with textual descriptions based on their distribution. This approach is inspired by the box plot method~\cite{tukey1977exploratory}, relating feature values to their median. Each numerical feature value in the input/output is labelled with one of five class labels:
\begin{itemize}
\item Lower outlier: values below the 0.1 percentile
\item Lower whisker: values around the 25th percentile
\item Median: at the 50th percentile
\item Upper whisker: values around the 75th percentile
\item Upper outlier: values above the 99.9 percentile.
\end{itemize}
For example, the text '\emph{petal length (cm): 30}' becomes '\emph{petal length (cm): 30 (lower whisker)}', which indicates that the value is between 0.1 percentile and the 25th percentile.

\paragraph{Relation encoding (RE)}
We have found out that substituting special signs with their textual descriptions improves the LLM result. This is consistent with observations in~\cite{dave2024investigating}. Therefore, to improve the LLM performance, mathematical signs for relations such as  ($>$,$=$,$<$) are substituted by their textual description, i.e. 'is less than', 'is equal to', 'is greater than'. 

\subsubsection{Input (prompt) conversion}
The input (prompt) template for every example $\vect{x}\in\set{X}$, can be described as:
\begin{equation*}
\underbrace{\text{prompt header}}_{\text{custom header text}}\quad
\underbrace{\text{ft}^1 \vect{x}^1 \ldots,\text{ft}^j \vect{x}^j \ldots, \text{ft}^F \vect{x}^F}_{\text{encoded features}}\quad
\underbrace{\text{prompt footer}}_{\text{custom footer text}},
\end{equation*}
where $\text{ft}^j$ is the name of the $j$th feature and $\vect{x}^j$ is the value of this feature in vector $\vect{x}$. 

The prompt header is always the same: 
\begin{itemize}
\item[] \emph{Here is the description of system state:}
\end{itemize}

The format of encoded features depends on preprocessing parameters described in Section~\ref{sec:method:preprocessing}, for example:
\begin{itemize}
    \item No preprocessing: \emph{sepal length (cm): 6.80, sepal width (cm): 2.80, petal length (cm): 4.80, petal width (cm): 1.40}
    \item IN$+$VD$+$RE: \emph{sepal length (cm): 69 (upper whisker), sepal width (cm): 33 (lower quantile), petal length (cm): 64 (upper quantile), petal width (cm): 54 (upper quantile)}
\end{itemize}

The prompt footer uses a single template, which is also modified to match the format of preprocessing parameters, e.g. for IN$+$VD$+$RE (all options enabled):
\begin{itemize}
\item[] \emph{"Based on values of system features, classify the state of the system and explain the decision. Use logical rules comparing feature values with thresholds using 'is greater/less than'. Format: '[feature\_name] [value] [inequality] [threshold]', for example: 'feature\_name 0 (lower whisker) is less than 10 (upper quantile). Label: 0'. Provide a classification label from {uniqe\_labels}. Explanation and system label:}
\end{itemize}

\subsubsection{LLM output prepration}
The expected textual LLM output for an input vector $\vect{x}\in\set{X}$ is generated by sampling the random forest classifier (RFC)~\cite{breiman2001random}. This involves randomly selecting a single decision tree classifier (DTC) from the RFC -- an ensemble of DTCs that label examples using majority voting -- and extracting the decision path for the example $\vect{x}$.

\paragraph{Decision path selection in DTC}
A decision tree classifier is a tree model where internal nodes represent conditional decisions based on feature values, and terminal nodes (leaves) provide class labels derived from these decisions. Given a set of classes $\set{Y}$, the possible classification outcomes form a set of decision paths. A decision path $\set{P}$ is defined as:
\begin{equation}
 \label{eq:decision_path}
 \set{P} = \{(n_1,n_2,\ldots,n_H), y\},
\end{equation} 
where the label $y\in\set{Y}$ is associated with the terminal node, the decisions at nodes in the path are denoted as $n_{k}$ and $H$ is the length of this path\footnote{We note that the $n_k$ are dependent on $\set{P}$, so should formally be defined as $n^\set{P}_k$, but we skip the path index for convenience.}. A decision in $k$-th node $n_k$ in the path is defined as a tuple:
\begin{equation}
\label{eq:node}
n_k=(j_k,t_k,d_k),    
\end{equation}
where $j_k<F$ is a feature index, $t_k\in\R$ is a feature threshold, and $d_k\in\{-1,+1\}$ is an indication of whether the right or left path has been selected in the tree, which translates into a conditional statement between the feature value and the threshold. 

We define an example $\vect{x} \in \set{X}$ as belonging to the path if:
\begin{equation}
\vect{x}\in\set{P} \iff \forall_{k=\{1,\ldots,H\}}{\begin{cases}
\vect{x}^{j_k}\in(-\infty,t_{k}) \text{ if } d_k\le0\\
\vect{x}^{j_k}\in(t_{k},\infty) \text{ otherwise,} 
\end{cases}}
\end{equation}
where $\vect{x}^{j_k}$ denotes the $j_k$-th element (a feature) of the vector $\vect{x}$. In other words, an example belongs to the path if it fulfils all of the path conditions.

\paragraph{Sampling the RFC}
Given a set of decision tree classifiers in the random forest classifier, for every labelled example  $(\vect{x}, y)\in\vect{T}_\text{train}$ we randomly select a single decision tree classifier and a decision path $\set{P}$ in this tree, such that $\vect{x}\in\set{P}$ and $y=y^\set{P}$ where $y^\set{P}$ is the label assigned by the path $\set{P}$ i.e. this decision path correctly classifies the example. We repeat this operation $n_\text{trees} \ge 1$ times, where $n_\text{trees}$ is a hyperparameter of the algorithm. Each sampling repetition generates a unique tuple $(\vect{x},y,\set{P})$, provided that it exists. The set of all returned tuples $\set{T}_\text{LLM}$ will be used to train LLM after the conversion of feature vectors and decision paths into textual forms. It is possible that the size $|\set{T}_\text{LLM}|>|\set{T}_\text{train}|$ if the value $n_\text{trees}>1$. The idea behind this extension is that the LLM may learn to generate explanations that combine multiple decision paths, possibly finding a better description of class features.

\paragraph{Output conversion} 
\label{sec:output_conversion}
Given an example $\vect{x}\in\set{X}$ and a feature index $j$, a statement given by the Eq.~\ref{eq:dtree_rule} can be created from a feature name $\text{ft}^j$, feature value $\vect{x}^j$, the condition and the threshold value $t$. Therefore, for a decision path $\set{P}$ defined by the Eq.~\eqref{eq:decision_path}, and an example $\vect{x}\in\set{P}$, any decision $n_k$ in this path, defined by the Eq.~\eqref{eq:node}, can be expressed in form of conditional statement given by the Eq.~\ref{eq:dtree_rule}. This is because the feature name $\text{ft}^j$ for any feature index $j$ is known, and the decision given by Eq.~\eqref{eq:node} contains the feature index $j_k$, the condition (as an indicator $d_k$) and the threshold value $t_k$. This is used to create the LLM output representing a decision path for any labelled example in the form:
\begin{equation}
\label{eq:output}
\underbrace{\text{statement}_1 \text{ and } \ldots \text{ and } \text{statement}_H}_{\text{Conditions in textual form}}\quad
\text{Label: }\underbrace{\text{class label}}_{\text{Label of the example}}
\end{equation}

These textual representations of the decision path undergo postprocessing described in Section~\ref{sec:method:preprocessing} depending on whether IN, VD, and RE steps are enabled, for example:
\begin{itemize}
    \item[] No preprocessing: \emph{petal length (cm) $4.80 < 4.85$ and petal width (cm) $1.40 > 0.80$. Label:~$1$}
    \item[] Preprocessing options IN$+$VD$+$RE: \emph{petal length (cm) 64 (upper quantile) is less than 64 (upper quantile) and petal width (cm) 54 (upper quantile) is greater than 29 (lower quantile). Label: $1$}
\end{itemize}

\paragraph{Random forest classifier parameters:} We used a standard scikit-learn~\cite{scikit-learn} implementation of RF with $n=100$ trees. Due to the simplicity of the datasets used and to limit the length of the generated output, we have limited max depth to $d_\text{max}=2$. For every input, we have selected $n_\text{trees}=2$ trees from every forest (almost doubling the size of the training set).

\subsubsection{LLM fine-tuning with LoRA}
In the experiments, we use the LLM sequence-to-sequence FLAN-T5-base model, based on the T5 model \cite{raffel2020exploring} and fine-tuned with the FLAN dataset \cite{longpre2023flan}. We selected FLAN-T5 due to its ability to achieve comparable results to larger models \cite{zhang2023instruction}, with efficient performance and minimal computational overhead. We also tested other architectures, such as FLAN-T5-XL and Llama3, with no better results.

The model was trained using parameter-efficient fine-tuning (PEFT) with LoRA (Low-Rank Adaptation) ~\cite{hu2021lora}, targeting the $(q, v)$ model components for the seq2seq task. Training and inference were conducted using the Transformers library \cite{wolf2020transformers}.

During training, the data (pairs of generated inputs/outputs) was split into stratified training, test, and validation sets, ensuring that all input-output pairs with similar inputs are in one set to avoid information leakage between training/test set if $n_\text{trees}>1$. 

\paragraph{Training parameters:}
We used the following LoRA configuration: rank $r=32$, scaling factor $alpha=32$, targeted model modules $(q, v)$, dropout $d=0.05$. The training was performed with learning rate $lr=10^{-3}$ and the number of epochs $n=150$. Output generation parameters: maximum new tokens $t_{max}=300$, number of beams $b_n=6$, beam groups $b_g=2$, diversity penalty $p_d=0.5$, temperature $t=0.5$.

\subsection{LLM output validation}
\label{sec:output_validation}
While objective assessment of classifier accuracy with a classification metric such as balanced accuracy~\cite{brodersen2010balanced} is straightforward, assessment of provided explanations is more complicated. While human evaluation~\cite{ziems2023explaining} is a valid option, it is also costly. Therefore, we propose to verify the correctness of the explanation based on the properties of the subset of the training set $\set{T}_\text{train}$ designated by this explanation. As explained in Section~\ref{sec:output_conversion}, for any training example $\vect{x}\in\set{X}_\text{train}$ the decision path $\set{P}$ of the RFC trained on the set $\set{T}_\text{train}$ can be represented in the textual form given by the Eq.~\ref{eq:output}. Therefore, for any example $\vect{x}\in\set{X}$, we expect that the output of the trained LLM will represent and can be parsed into an LLM-generated decision path $\set{P}_\text{LLM}$. We claim that conditions (as defined by the Eq.~\eqref{eq:node}) in this path explain its label -- the subset of the training set determined by these conditions belongs to the class assigned by the path $\set{P}_\text{LLM}$. 

To verify the correctness of this claim we find the subset $\set{T}_\text{predicted}\subset\set{T}_\text{train}$, such that for every $\vect{x}\in\set{T}_\text{predicted} \iff \vect{x}\in\set{P}_\text{LLM}$ and expect that this subset contains mostly examples of the class $y\in{\set{P}_\text{LLM}}$, assigned by the LLM. We verify the correctness of LLM explanation using precision and recall metrics applied to the set $\set{T}_\text{predicted}$, using the predicted label from LLM output and true labels from the training set.

The validation is performed by parsing the rules with a parser that transforms the LLM output into Python code. We use a dedicated regular expression-based parser, but alternatively, this task can be performed by querying a large language model (ChatGPT 3.5) using the following prompt template (modified, depending on preprocessing described in Section~\ref{sec:method:preprocessing}):

\emph{Translate decision tree rules "feature (value) operator threshold" directly into Python pandas code to filter features\_df. 
The output should strictly be Python code, consisting of filtering commands only. For "petal width <= 0.80", provide filtered\_df = filtered\_df[filtered\_df["petal width"] <= 0.8]. Apply each rule to filtered\_df sequentially. No additional text or explanation is needed. Text to translate:}

As a result, example text:

\emph{petal length (cm) 6.70 > 2.45 and sepal length (cm) 6.70 > 6.75. Label: 1}

is converted into code:
\begin{Verbatim}[xleftmargin=1cm]
filtered_df = filtered_df[filtered_df['petal length (cm)'] > 2.45]
filtered_df = filtered_df[filtered_df['sepal length (cm)'] > 6.75]
\end{Verbatim}

This code is applied to the data frame and a label vector in the set $\set{T}_\text{train}$ to compute performance metrics.

%\subsection{Reference models}
%do we use reference? Maybe not?

\subsection{Datasets}
We are using three well-known datasets: iris, wine, and breast cancer.

\section{Results}
Results are presented in Table~\ref{tab:res_preprocessing} for experiments with relation encoding (RE) enabled and Table~\ref{tab:res_nopreprocessing} for RE disabled. These tables show label accuracy, mean statement accuracy, and recall from assessing explanation correctness as detailed in Section~\ref{sec:output_validation}. These metrics describe the average properties of the subset $\set{T}_\text{predicted}\subset\set{T}_\text{train}$ indicated by LLM explanations. Experiments were repeated five times with different random seeds, and the results were averaged.

The model's label accuracy was consistently high across all datasets, usually above 95\%. The results are comparable to those of the random forest classifier used to train the LLM, which achieved accuracies of $94.93\%$ (Iris), $97.30\%$ (Wine), and $94.72\%$ (Breast cancer) in a 5-fold CV experiment repeated five times (5cv5), consistent with findings in~\cite{dinh2022lift}.

For LLM explanations, we observed high average statement accuracy and recall with RE+IN+VN preprocessing, both above $80\%$ and $74\%$, respectively. This indicates that the LLM's statements accurately represent the model's assigned class.

Enabling RE sharply increased the number of correct (parsable) statements. Without RE, common errors included missing relation (e.g., inequality) signs, which were rare with RE enabled. This error pattern was consistent across multiple generations, regardless of generation parameters (such as temperature).
Table~\ref{tab:res_preprocessing} shows that RE+IN+VN preprocessing generally improves label and statement accuracy. However, the impact on statement recall is inconsistent, suggesting a potential trade-off between the accuracy and recall of generated statements.

\begin{table}
\begin{tabular}{llrrrr}
\toprule
 & DS name & Label accuracy & Statement accuracy & Statement recall & Correct \\
\midrule
0 & iris & 93.10 & 88.30 & 88.29 & 100.00 \\
1 & iris(IN) & 92.36 & 88.02 & 87.17 & 100.00 \\
2 & iris(VN) & 93.79 & 90.02 & 89.54 & 100.00 \\
3 & iris(IN+VN) & 94.81 & 90.27 & 89.18 & 100.00 \\
4 & wine & 91.67 & 80.54 & 77.11 & 100.00 \\
5 & wine(IN) & 98.33 & 77.11 & 70.67 & 99.44 \\
6 & wine(VN) & 94.44 & 85.39 & 80.92 & 100.00 \\
7 & wine(IN+VN) & 100.00 & 81.21 & 74.54 & 100.00 \\
8 & breast & 94.74 & 90.27 & 79.27 & 99.82 \\
9 & breast(IN) & 97.19 & 91.50 & 77.30 & 99.82 \\
10 & breast(VN) & 96.67 & 91.20 & 79.54 & 100.00 \\
11 & breast(IN+VN) & 97.02 & 91.14 & 77.25 & 99.82 \\
\bottomrule
\end{tabular}
\caption{Mean performance of LLM with relation encoding (RE) enabled. (IN) denotes integer normalisation, and (VD) denotes verbal description of values as described in Section~\ref{sec:method:preprocessing}. Label accuracy is the accuracy based on assigned labels. Statement accuracy and recall are computed based on a subset of training data indicated by LLM-generated statements as described in Section~\ref{sec:output_validation}. `Correct' denotes the percentage of statements that could be properly parsed. }
\label{tab:res_preprocessing}
\end{table}

\begin{table}
\begin{tabular}{llrrrr}
\toprule
 & DS name & Label accuracy & Statement accuracy & Statement recall & Correct\\
\midrule
0 & iris & 93.10 & 78.93 & 78.77 & 37.33\\
1 & iris(IN) & 94.12 & 81.89 & 82.07 & 35.10\\
2 & iris(VN) & 93.10 & 80.70 & 81.22 & 37.33\\
3 & iris(IN+VN) & 94.45 & 83.80 & 84.78 & 34.05\\
4 & wine & 88.89 & 70.41 & 79.19 & 37.78\\
5 & wine(IN) & 93.89 & 74.90 & 83.06 & 36.11\\
6 & wine(VN) & 97.22 & 85.61 & 92.44 & 31.11\\
7 & wine(IN+VN) & 97.22 & 78.09 & 86.07 & 31.11\\
8 & breast & 93.68 & 91.00 & 73.48 & 33.51\\
9 & breast(IN) & 96.14 & 95.07 & 74.84 & 31.58\\
10 & breast(VN) & 96.27 & 93.28 & 76.63 & 32.46\\
11 & breast(IN+VN) & 98.25 & 95.25 & 75.26 & 31.80\\
\bottomrule
\end{tabular}
\caption{Mean performance of LLM without relation encoding (RE) enabled. (IN) denotes integer normalisation, and (VD) denotes verbal description of values as described in Section~\ref{sec:method:preprocessing}. Label accuracy is the accuracy based on assigned labels. Statement accuracy and recall are computed based on a subset of training data indicated by LLM-generated statements as described in Section~\ref{sec:output_validation}. `Correct' denotes the percentage of statements that could be properly parsed.}
\label{tab:res_nopreprocessing}
\end{table}

\subsection{Discussion}
We observed that changes in LLM training hyperparameters moderately impact performance. The most critical parameter is the depth of the RF, as more complex datasets require deeper decision trees, resulting in more extended outputs. Analysing the results, we suspect that improved performance may be achieved with better (non-random) selection of decision paths and reduction of redundant conditions. The number of training epochs used in our experiment was sufficient for the tested datasets, but larger datasets may require more extended training.

We have also applied the described methodology to the hyperspectral classification problem~\cite{romaszewski2016semi} using standard datasets like the `Pavia University' image. Initial experiments in spectral classification achieved 78\% accuracy (similar parameters as in the paper) to 85\% accuracy in spectral classification, depending on the RF parameters, indicating the potential for applying this methodology to Computer Vision problems.

Initial experiments with different models, such as FLAN-T5-XL, showed results comparable to FLAN-T5-base, likely due to the simplicity of the training sets used. Experiments with GPT-3.5 and LLaMA models initially showed worse performance, possibly due to limited training time for these larger models. 
We have also tried reformulating inputs and outputs using GPT-4. The model was instructed to reduce redundant statements, paraphrase the text while maintaining meaning and information, and create different text versions. After fine-tuning with such preprocessed data, the LLM generated text resembling paraphrased rules in natural language. However, verification of whether such paraphrasing does not degrade classification performance needs additional experiments. This capability may be a crucial step toward developing a model with chatbot functionality, which could function as a class-aware expert.
 
\section{Conclusions}
Our results demonstrate that an LLM trained with an RF classifier can accurately classify and explain its decisions. The high classification accuracy aligns with recent works such as~\cite{dinh2022lift}, and the quality of explanations is promising, as the LLM effectively segments data space fragments where class examples are concentrated. Additionally, our proposed statement verification method is less costly than human-assisted assessment~\cite{zhuang2024learning} and may be a promising approach for generating data for reward model creation \cite{ouyang2022training}.

We believe the proposed training method can be applied in many applications and is particularly suited for recommender systems \cite{hou2024large}. LLMs fine-tuned to understand the structure of numerical data can explain or support an ensemble of classical models.

\section{Acknowledgement}
The authors would like to thank Prof. Maciej Piasecki and Prof. Julian Szymański for their valuable comments during the PP-RAI'24 conference, particularly regarding LLMs' understanding of numerical data in the context of the tokenization mechanism.

\bibliographystyle{plainnat}
\bibliography{ttt_llm}

\end{document}